\documentclass[letterpaper, 10pt, conference]{ieeeconf}
\IEEEoverridecommandlockouts
\usepackage{amsmath}
\usepackage[linesnumbered,lined,boxed,commentsnumbered]{algorithm2e}
\SetAlCapSkip{0.1cm}
\SetAlgoCaptionSeparator{.}
\usepackage{graphicx}
\usepackage{float}
\usepackage{siunitx}
\usepackage{cite}
\usepackage{color}
\usepackage{amssymb}

\usepackage[dvipsnames]{xcolor} 
\usepackage[commentmarkup=uwave]{changes} 
\definechangesauthor[name={kaiwen}, color=Maroon]{kaiwen}

\definecolor{RightColor}{RGB}{39, 174, 96} 
\definecolor{WrongColor}{RGB}{202, 62, 71} 
\definecolor{QueryColor}{RGB}{155, 89, 182} 
\makeatletter
\let\NAT@parse\undefined
\makeatother
\usepackage[colorlinks, linkcolor=black, anchorcolor=black, citecolor=black]{hyperref}

\usepackage{multirow}
\usepackage{tablefootnote}
\usepackage{threeparttable}
\usepackage{booktabs}
\graphicspath{ {./figures/} }

\usepackage{epstopdf}
\usepackage{makecell}
\newcommand{\sysname}{\texttt{AutoPlace}\xspace}

\begin{document}
\title{AutoPlace: Robust Place Recognition with Single-chip Automotive Radar}


\author{Kaiwen Cai\authorrefmark{2}, Bing Wang\authorrefmark{4}, Chris Xiaoxuan Lu\authorrefmark{3}\authorrefmark{1}\\
\authorrefmark{2}University of Liverpool,
\authorrefmark{4}University of Oxford,
\authorrefmark{3}University of Edinburgh
\thanks{\authorrefmark{1}Corresponding author: Chris Xiaoxuan Lu (xiaoxuan.lu@ed.ac.uk)}
}


\maketitle

\begin{abstract}
 This paper presents a novel place recognition approach to autonomous vehicles by using low-cost, single-chip automotive radar. Aimed at improving recognition robustness and fully exploiting the rich information provided by this emerging automotive radar, our approach follows a principled pipeline that comprises (1) dynamic points removal from instant Doppler measurement, (2) spatial-temporal feature embedding on radar point clouds, and (3) retrieved candidates refinement from Radar Cross Section measurement. Extensive experimental results on the public nuScenes dataset demonstrate that existing visual/LiDAR/spinning radar place recognition approaches are less suitable for single-chip automotive radar. In contrast, our purpose-built approach for automotive radar consistently outperforms a variety of baseline methods via a comprehensive set of metrics, providing insights into the efficacy when used in a realistic system.

\end{abstract}

\section{Introduction}

By recognizing revisited places when traveling, place recognition is a key enabler to mobile autonomy and plays essential roles in a wide range of downstream tasks such as scene understanding, loop closure detection, localization, etc.

\begin{figure}[t]
    \centering
    \includegraphics[width=3.45in]{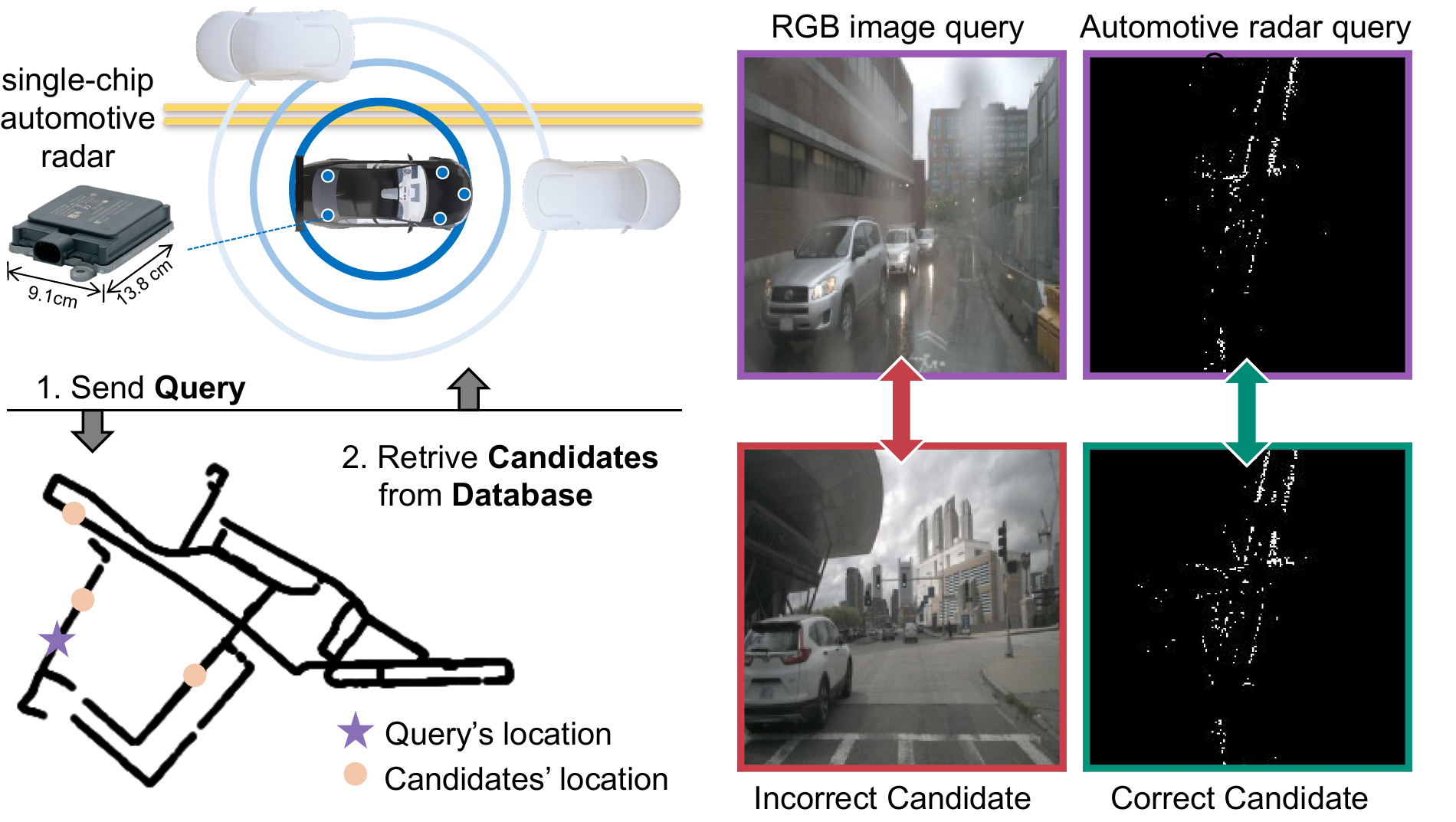}
    \caption{Place recognition using the single-chip automotive radar: given a query (marked in \textcolor{QueryColor}{purple}) acquired from the same place on a rainy day~\cite{caesarNuScenesMultimodalDataset2020},
    the state-of-the-art RGB camera-based place recognition \cite{arandjelovicNetVLADCNNArchitecture2016} failed to retrieve the correct candidate due to raindrops blocking the camera, while the proposed \sysname successfully retrived the correct one. 
    }
    \label{openfig}
\vspace{-0.5cm}
\end{figure}

Visual place recognition develops with the prevalence of commercial RGB cameras, which involves handcrafted features\cite{cumminsFABMAPProbabilisticLocalization2008, milfordSeqSLAMVisualRoutebased2012} and deep learning-based ones\cite{sunderhaufPerformanceConvNetFeatures2015,arandjelovicNetVLADCNNArchitecture2016,zhuAttentionbasedPyramidAggregation2018}. On the other hand, a LiDAR sensor is usually adopted as an alternative optical sensor for place recognition due to its better robustness in dim and dark environments. While place recognition approaches based on these optical sensors have made considerable progress in the past decade\cite{arandjelovicNetVLADCNNArchitecture2016, zhuAttentionbasedPyramidAggregation2018, heM2DPNovel3D2016, uyPointNetVLADDeepPoint2018, komorowskiMinkLoc3DPointCloud2020}, they still fall short under visual degradation common on the roads (e.g., rain, snow, dust, fog, and direct sunlight). An example is shown in Fig. \ref{openfig}, where the state-of-the-art RGB camera-based place recognition failed to retrieve the correct candidate due to raindrops blocking the camera.

Unlike the above optical sensors operating in the visible spectrum, radar operates in a millimeter-wave frequency band, lending itself a modality robust to scene illumination and airborne obstacles\cite{luMilliEgoSinglechipMmWave2020}. By taking input as the dense radar spectra/images, recent works\cite{saftescuKidnappedRadarTopological2020,barnesRadarLearningPredict2020} have shown the feasibility of place recognition based on \emph{mechanically spinning radar} (e.g., CTS-350X). Despite the impressive performance achieved, spinning radar is known to be bulky and costly\cite{schumann2021radarscenes} and require to be mounted on the roof of the vehicle, nor able to provide the Doppler information. In contrast, \emph{automotive radar} (aka. single-chip millimeter-wave radar) emerge as a low-cost and lightweight alternative that is pervasively embraced 
by major vehicle manufacturers (e.g., Audi and Ford \cite{audi, ford}). As it adopts wave beamforming rather than mechanical spinning to scan the environment\cite{luSeeSmokeRobust2020}, an automotive radar can measure the point's radial velocity through the Doppler effect as well as fine-grained Radar Cross Section (RCS). 

These unique characteristics make automotive radar an attractive sensor for autonomous driving. While prior works use them to estimate the vehicle's (ego-)motion\cite{kungNormalDistributionTransformBased2021,linDepthEstimationMonocular2020} and localization~\cite{rapp2015feature, schuster2016landmark}, realizing robust place recognition by automotive radar has never been explored and features different challenges. Compared with spinning radars \cite{saftescuKidnappedRadarTopological2020}, 
automotive radars' point clouds are significantly noisier, sparser and in much lower resolution~\cite{luMilliEgoSinglechipMmWave2020}. These low-quality point clouds results in unreliable global descriptors that often `mislead' a place recognition system.

In this work, we exploit the unique characteristics of automotive radar and propose a robust \underline{Auto}motive radar \underline{Place} recognition approach dubbed \sysname to address the above challenges.
Specifically, our contributions are:
\begin{itemize}
\item This paper is the first work that validates the capability of single-chip automotive radar for place recognition.
\item We propose a novel place recognition method by fully utilizing the radial velocity and RCS measurement and effectively modeling the spatial and temporal radar points with a compact deep neural network.
\item The proposed \sysname consistently outperforms a variety of competing approaches on the public nuScenes dataset \cite{caesarNuScenesMultimodalDataset2020}, with code avaliable at: \url{https://github.com/ramdrop/AutoPlace}. 
\end{itemize}

\begin{figure}[t]
    \vspace{0.15cm}
    \centering
    \includegraphics[width=3.4in]{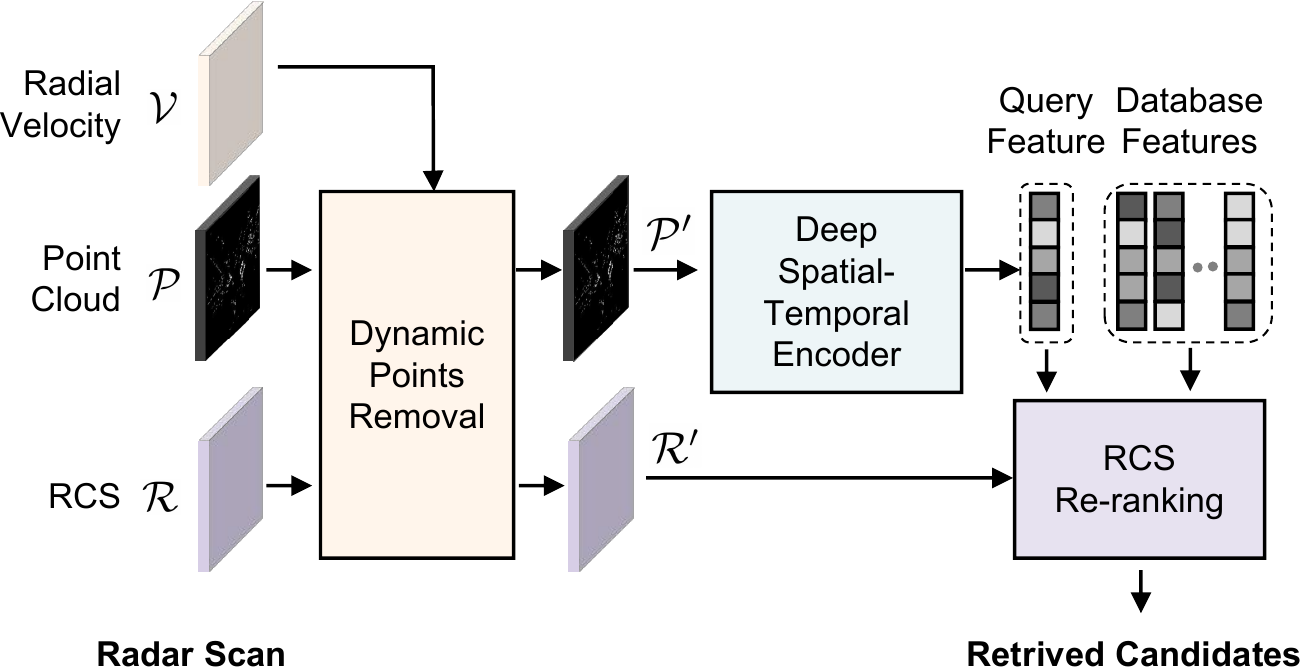}
    \caption{Overview of \sysname: radial velocity, point cloud and RCS measurement of a radar scan are utilized by the proposed \textit{Dynamic Points Removal} method, \textit{Deep Spatial-Temporal Encoder} and \textit{RCS Histogram Re-ranking} method, respectively.}
    \label{pipeline}
\vspace{-0.5cm}
\end{figure}

\section{Related Work}

Place recognition with cameras is an established topic due to the sensor ubiquity, rich information and cost-effectiveness. Early works use handcrafted features and heuristic sequence matching methods\cite{cumminsFABMAPProbabilisticLocalization2008, milfordSeqSLAMVisualRoutebased2012}, while Convolution Neural Networks (CNNs) and attention mechanisms\cite{arandjelovicNetVLADCNNArchitecture2016, zhuAttentionbasedPyramidAggregation2018} are recently explored. Unlike cameras, LiDAR sensors measure objects with explicit scales. LiDAR place recognition approaches include spatial segments methods\cite{heM2DPNovel3D2016, kimScanContextEgocentric2018}, point-wise neural networks\cite{uyPointNetVLADDeepPoint2018} and 3D convolution networks\cite{komorowskiMinkLoc3DPointCloud2020}. 

Once as a common sensor used on ships \cite{RadarRobotCarDatasetICRA2020}, mechanically spinning radar has recently been utilized for vehicle place recognition. UnderTheRadar\cite{barnes2020under} uses intermediate features as global decsriptor for place recognition. RadarSLAM\cite{hongRadarSLAMRadarBased2020} uses M2DP\cite{heM2DPNovel3D2016} to generate compact descriptors for radar point clouds. KidnappedRadar\cite{saftescuKidnappedRadarTopological2020} exploits a variant of NetVLAD\cite{arandjelovicNetVLADCNNArchitecture2016} as the feature extractor with sophisticated modification to improve rotational invariance.


The closest work to ours is LookAroundYou\cite{gaddLookYouSequencebased2020}, which improves KidnappedRadar\cite{saftescuKidnappedRadarTopological2020} by using the off-the-shelf sequence matching mechanism from SeqSLAM\cite{milfordSeqSLAMVisualRoutebased2012} . However, LookAroundYou\cite{gaddLookYouSequencebased2020} inevitably inherits the limitations of SeqSLAM\cite{milfordSeqSLAMVisualRoutebased2012} in the following aspects\cite{chancanDeepSeqSLAMTrainableCNN2020}: (1) It requires at least two complete trajectories - query trajectory and database trajectory, and they should be aligned with the same number of samples, while our method can work with the queries and databases of any size, and has no dependency on fine-grained sequence alignments. (2) It requires the whole trajectory to be spatially continuous for performing the local contrast enhancement\cite{milfordSeqSLAMVisualRoutebased2012}, while our method only requires a locally continuous sequence. (3) It heuristically matches sequences, while our method end-to-end trains a sequential matching network. More importantly, rather than uses the bulky spinning radar, our work goes for lightweight automotive radar, by which richer measurements are provided and used for more robust place recognition.



\section{Methods}

Fig. \ref{pipeline} illustrates the pipeline of the proposed \sysname. An automotive radar scan provides a set of measurements, including the \emph{point cloud}
$\mathcal{P}=\{ (x_i, y_i)|i=1,2,..,N\}$, \emph{radial velocity} $\mathcal{V}=\{ v_i|i=1,2,..,N\}$ and \emph{RCS} $\mathcal{R}=\{ r_i|i=1,2,..,N\}$, where $N$ is the number of observed points at a scanning instant. When receiving this input, our proposed Dynamic Points Removal (DPR) module first exploits $\mathcal{V}$ to identify and remove dynamic points in $\mathcal{P}$ and $\mathcal{R}$, and produces the refined  $\mathcal{P^\prime}$ and $\mathcal{R^\prime}$. Then a Deep Spatial-Temporal Encoder extracts a discriminative feature vector from $\mathcal{P^\prime}$. Finally, during the inference phase, our proposed RCS Histogram Re-ranking (RCSHR) module re-ranks the list of best matching candidates. We now detail the design of each module in what follows.

\subsection{Dynamic Points Removal}
\label{ssub:dpr}


It is common that dynamic and stationary objects coexist on a road, such as vehicles, pedestrians, traffic cones and building walls. While static objects are temporally consistent, dynamic ones are not: a moving vehicle may disappear when the vehicle revisits the same place. Consequently, dynamic objects could mislead the place recognition due to landmark inconsistency. To address this challenge, we propose a novel \emph{Dynamic Point Removal} method to generate a dynamic points mask $\mathcal{M}$ considering the two
motion status of the ego-vehicle (i.e., the status of the radar sensor):

\subsubsection{Moving Ego-vehicle}
\label{DTR_RANSAC}
Intuitively, when an ego-vehicle moves, the stationary objects move towards the opposite direction in ego-vehicle's field of view. In this case, the identification of these moving objects should be straightforward by using automotive radar because objects' velocities can be directly measured and returned from such sensors. Nevertheless, automotive radar can only provide the radial velocity for an observed object rather than the full velocity. The measured velocity from the radar can thus be "zero" when an object moves tangentially to the radar. To address this ambiguity of velocity measurement, we propose to distinguish the dynamic points based on radial velocities of all points in a scan rather than that of a single point. 

Formally, the points' radial velocity for stationary objects has a sinusoidal function as follows: 
\begin{equation}
    \label{v_profile}
v_{r,i} = -v_s\cos{(\alpha-\theta_{i})}, i=1,2,...,N
\end{equation}
where $v_s$ is the radar velocity, $\alpha$ heading direction, $N$ the number of points, $v_{r,i}$ and $\theta_{i}$ the radial velocity and azimuth angle of the $i^{th}$ point. Recall that point's radial velocity for moving object does not depend on the ego-vehicle motion and can be arbitrary, which means it does not necessarily fit the Eq. (\ref{v_profile}) and thus is a outlier. Following \cite{kellnerInstantaneousEgomotionEstimation2013}, we use the Least Square approach and the Random Sample Consensus (RANSAC)\cite{fischler1981random} algorithm to solve $v_s$ and $\alpha$ and identify outliers, by which the moving points are found.



\subsubsection{Static Ego-vehicle}
One limitation of the above dynamic points removal approach is that it works on the assumption that the ego-vehicle is moving such that 
the Eq.(\ref{v_profile}) can be utilized to identify outliers.
However, such an assumption does not hold when the ego-vehicle is static. To address this problem, we propose a simple yet effective method to identify dynamic points when the ego-vehicle is static. The key intuition behind our method is: at a scanning instant, if most points have zero radial velocities, then the ego-vehicle is very likely to be static. Furthermore, as a nonzero radial velocity can only result from a nonzero velocity, it is reasonable to regard points with nonzero radial velocity as moving points. The implementation flow of our method is described in Algorithm \ref{ag_dpr}.

\begin{figure}[t]
    \centering
    \includegraphics[width=3.5in]{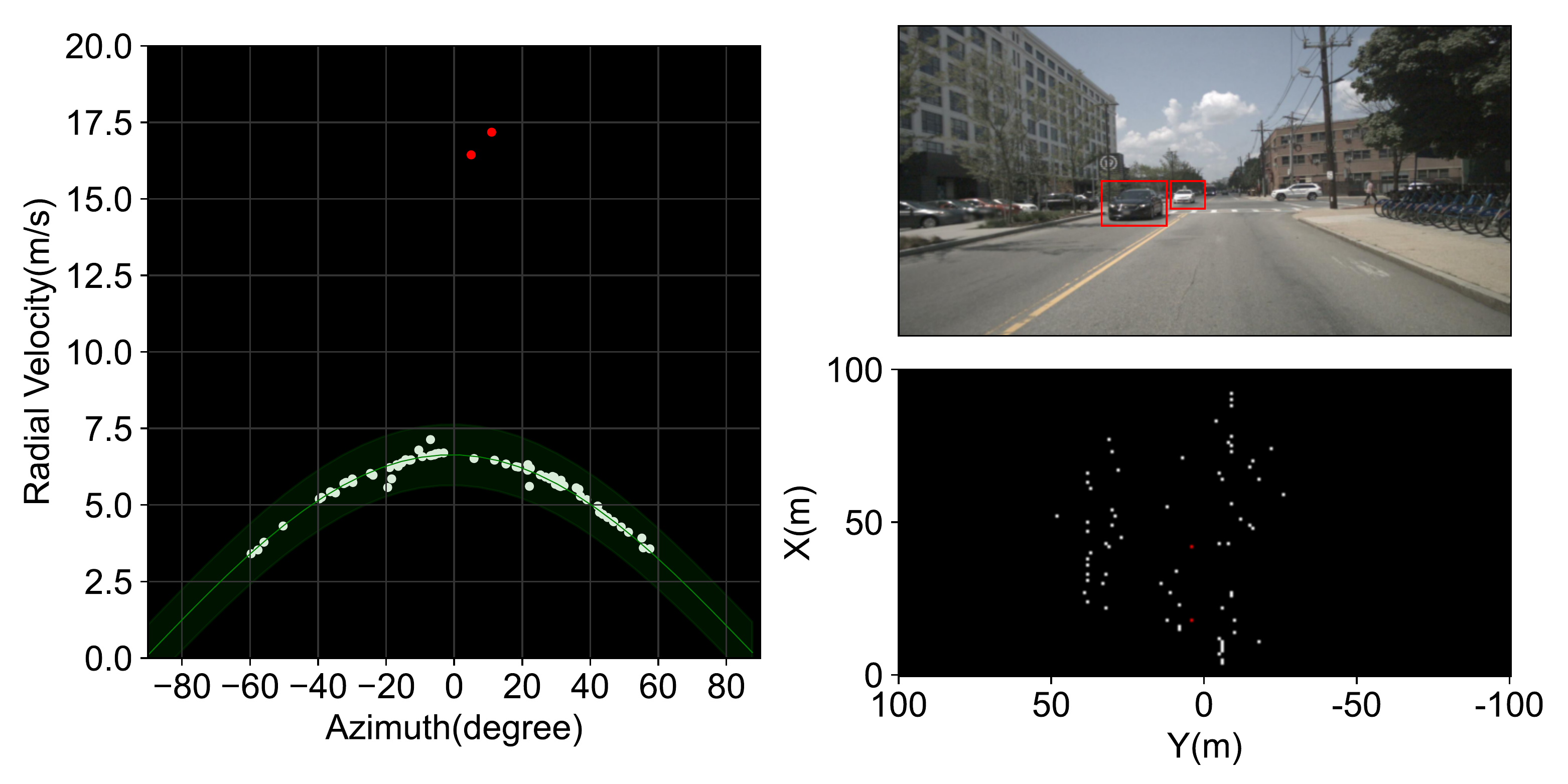}
    \setlength{\abovecaptionskip}{-0.6cm}
    \caption{Remove dynamic points based on radial velocity: the left figure shows the points' radial velocity distribution measured by the front radar in a single radar scan, and the right is the front-view images from the front camera, the bird view radar point cloud from the front radar, respectively. The two dynamic points (the moving cars) are successfully identified by the proposed DPR method and marked in \textcolor{red}{red}.}
    \label{DTR}
\vspace{-0.1cm}
\end{figure}

Fig. \ref{DTR} shows an example of points' radial velocity distribution, the corresponding bird view radar point cloud, and the identified dynamic points using the proposed DPR method\footnote{More illustrative examples can be found on our website: https://github.com/ramdrop/autoplace.
}.

\vspace{-0.4cm}
\setlength{\textfloatsep}{5pt}
\begin{algorithm}[h]
    \LinesNotNumbered
    \label{ag_dpr}
    \SetKwInOut{Input}{Input}\SetKwInOut{Output}{Output}
    \caption{Dynamic Points Removal.}
    \Input{radial velocity  \( \mathcal{V}=\{ v_{r,i}, \theta _i | i=1,2..,N)\} \)\\veolcity fitting threshold \(\tau\)
    \\static velocity threshold \(v_r^{\tau}\)
    \\percentage of static points threshold \(p^\tau\)}
    \Output{dynamic points mask \(\mathcal{M} =\{d_i|d_i\in\{0,1\}, i =1,2,..,N\}\) (value $0$ means dynamic points)}
    
    \BlankLine
    Initialization: $p \leftarrow 0$,  \lForEach(){$d_i$}{$d_i \leftarrow 1$}
    \BlankLine
    \For{i =1 \KwTo N}{
        \lIf(){\(v_{r,i} < v_r^{\tau}\)}{\(p \leftarrow p + \frac{1}{N}\)}
        \lElse{\(d_i \leftarrow 0 \)}    
    }

    \texttt{/* static ego-vehicle */}

    \lIf(){\(p > p^\tau\)}{output $\mathcal{M}$   }
    \texttt{/* moving ego-vehicle */}

    \Else{
        \lForEach(){$d_i$}{$d_i \leftarrow 1$}
        execute RANSAC and Least Square algorithms to find outliers indexed as $j_1, j_2,.., j_n$\;
        \lForEach(){$j$ \rm in $\{j_1, j_2,.., j_n\}$}{$d_j \leftarrow 0$}
        output $\mathcal{M}$\;
    }    

\end{algorithm}
\vspace{-0.8cm}

\subsection{Deep Spatial-Temporal Encoder}
Once the dynamic points mask $\mathcal{M}$ is derived using the above DPR, refined $\mathcal{P^\prime}$ and $\mathcal{R^\prime}$ are then obtained by removing dynamic points in $\mathcal{P}$ and $\mathcal{R}$. 
Next, we 
introduce the \emph{Deep Spatial-Temporal Encoder} to encode radar point clouds spatial and temporal information for robust place recognition.

\label{network}
\begin{figure}[t]
    \vspace{0.15cm}
    \centering
    \setlength{\abovecaptionskip}{0.cm}
    \includegraphics[width=2.7in]{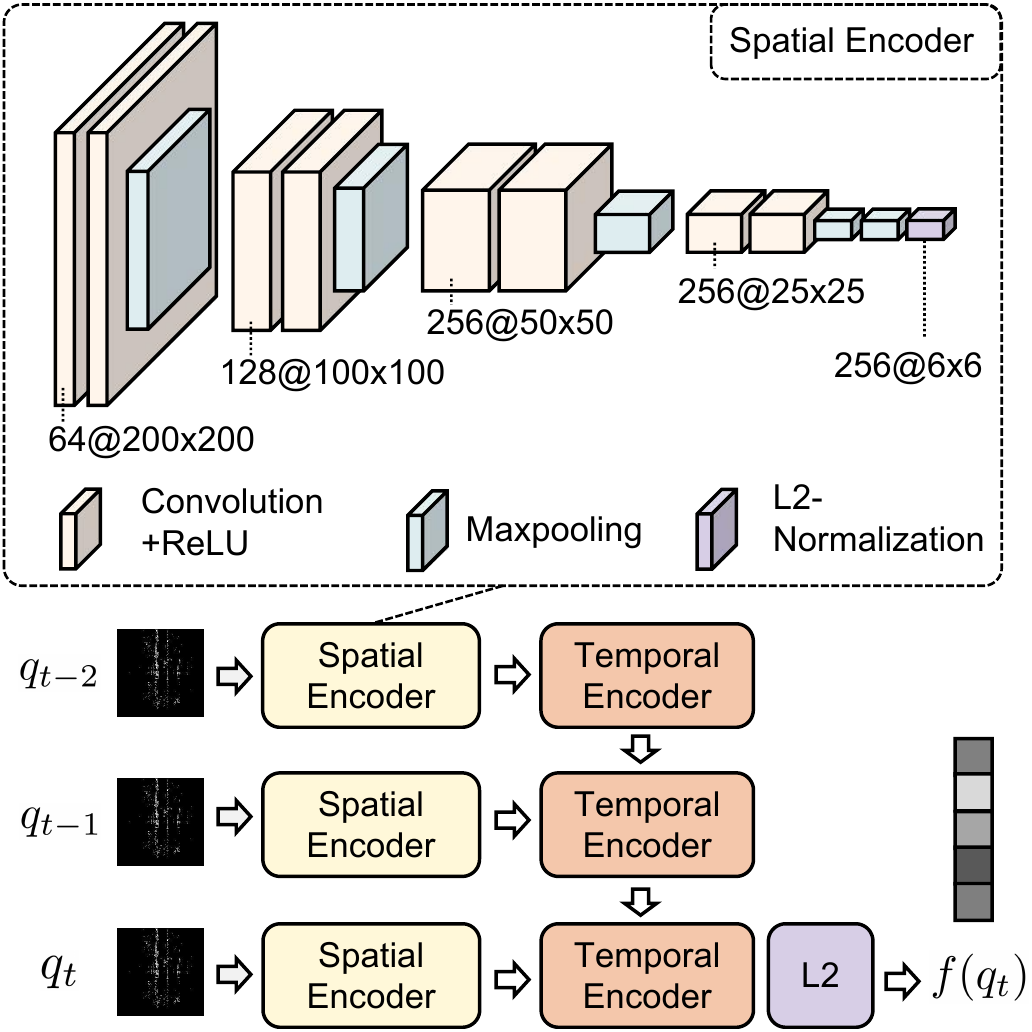}
    \caption{Overview of the proposed deep spatial-temporal encoder: in the Spatial Encoder diagram, the block color reflects layer type while the block size indicates the layer's output size, which is annotated as channel@width$\times$height. }
    \label{model}
\vspace{-0.1cm}
\end{figure}

As the automotive radar in this study provides 2D point clouds, we convert a radar point cloud to a \emph{radar image} by projecting all 2D points to the image panel with the occupied pixel assigned value $1$, and the pixels that are not occupied by any points are assigned value $0$. 



\subsubsection{Spatial Encoder}
\label{ssub:se}

To encode the spatial information of a radar image, we propose a convolution-based \emph{spatial encoder} as shown in Fig. \ref{model}: all convolutional layers have the same kernel size $(3 \times 3)$ and stride $(1 \times 1)$, followed by a ReLU layer. Max-pooling layers are used to downsample feature map, and all have a kernel size $(2 \times 2)$ and stride $(2 \times 2)$, except the last one has a larger kernel size $(3 \times 3)$ and a stride $(3 \times 3)$. An L2-normalization layer is used at the end of the spatial encoder. The feature map of the last layer has a size of $C \times H \times W$, which can be regarded as  $C$-dimensional features in $(H \times W)$ spatial locations.

\subsubsection{Temporal Encoder}
\label{ssub:te}

As point clouds given by automotive radar are noisy and sparse\cite{luMilliEgoSinglechipMmWave2020}, a measured object may come and go at random across consecutive radar images. Such inconsistency, however, makes it hard for the network to learn a consistent feature map even for the same place. 
To address this, we propose to utilize the temporal information in a series of radar images. The underlying idea here is that the object's inconsistency between consecutive scans can be mitigated by sequential smoothing. Inspired by the recent success of recurrent neural networks in smoothing sequential data\cite{wang2019recurrent, luMilliEgoSinglechipMmWave2020}, we here adopt the single-layer LSTM\cite{hochreiter1997long} as the \emph{temporal encoder} subsequent to the aforementioned spatial encoder. Fig. \ref{model} illustrates the overall architecture of the proposed deep spatial-temporal encoder.

\subsubsection{Loss Function}
Similar to NetVLAD\cite{arandjelovicNetVLADCNNArchitecture2016}, the triplet margin loss is used to train the model, which is given by
\vspace{-0.15cm}
\begin{equation*}
L=\sum\limits_{k} \max\{ d_E(f(q), f(p)) - d_E(f(q), f(n^k))+m, 0\}
\vspace{-0.15cm}
\end{equation*}
where $f(\cdot)$ denotes the network mapping a radar image to a feature vector, $d_E(\cdot)$ Euclidean distance, $q$ the query sample, $p$ the best positive matching sample, $n^k$ the true negative samples, $m=0.1$ the predefined margin, and $k=10$ the number of negative samples. See NetVLAD\cite{arandjelovicNetVLADCNNArchitecture2016} for more details about the triplet loss function.

\subsection{RCS Histogram Re-ranking}
\label{subs:RCSHR}

\begin{figure}[t]
    \centering
    \includegraphics[width=3.3in]{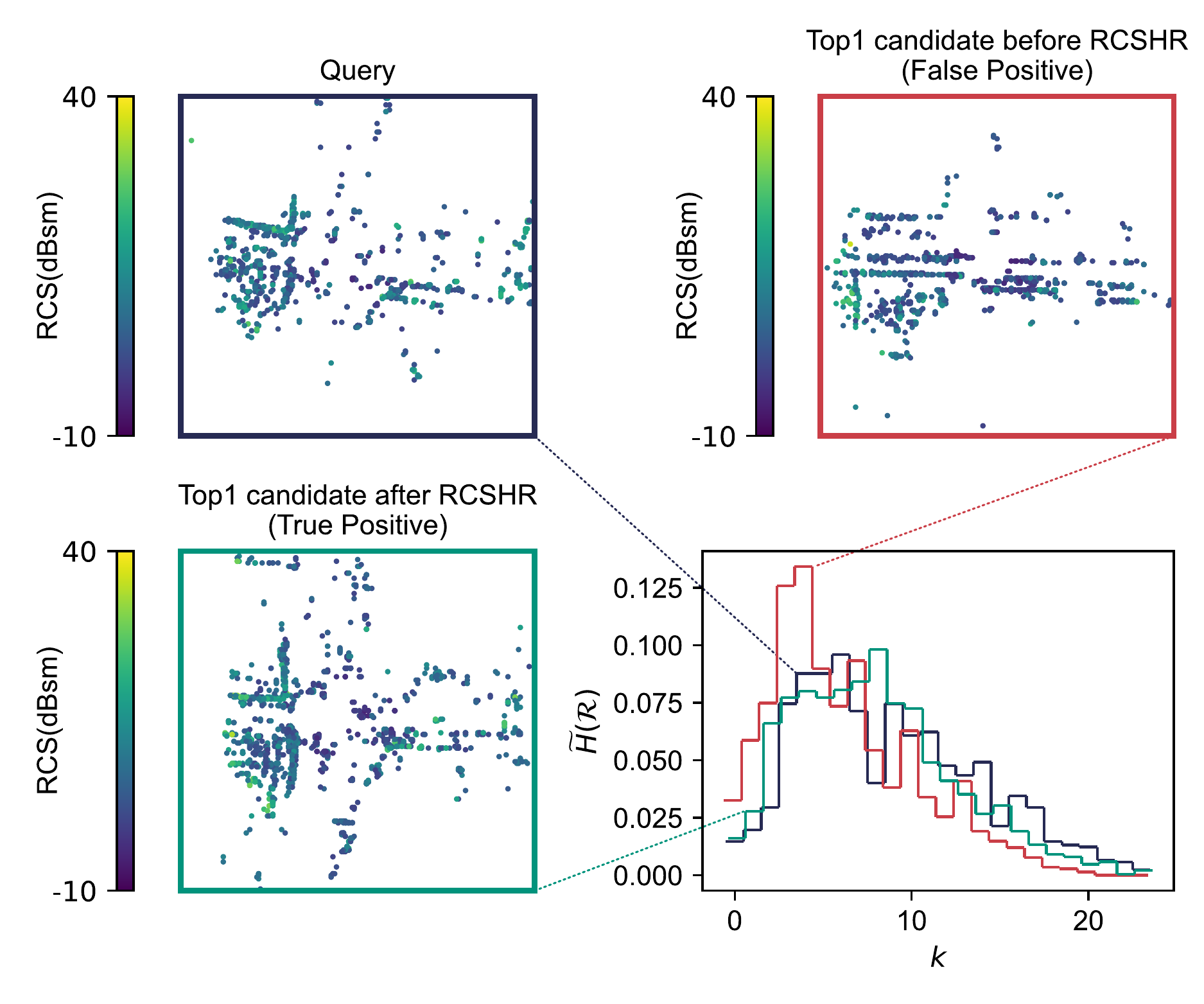}
    \setlength{\abovecaptionskip}{-0.3cm}
    \caption{An example of RCSHR: The lower right figure is the RCS histograms of a query and two candidates, while others are their bird view radar images. Given a query, the retrieved top-1 candidate before RCSHR is a false positive; after performing RCSHR based on their RCS histograms, we identified the true positive candidate. (we use point size larger than $1$ pixel for better visualization) }
    \label{RCSHR}
\vspace{-0.1cm}
\end{figure}

RCS values, as a type of radar output, are finally used in \sysname to refine the recognition accuracy. 
RCS is a property of the object's reflectivity, mainly determined by the object's material, size and reflected angle. This measurement provides an additional feature for an object and has proven effective in assisting odometry task \cite{joseAugmentedStateSLAM2005,liMillimeterWaveRadarSLAM2020}. However, the question remains whether and how the RCS can contribute to a place recognition task. To answer this question, we propose a novel method, dubbed \emph{RCS Histogram Re-Ranking} to further improve place recognition accuracy. The intuition is that geometrically-close places should share similar RCS histograms regardless of weather and illumination variances.
Therefore, when measuring the similarity of two radar point clouds, we consider not only their feature distance but also RCS histogram distance. Concretely, the steps of deriving the RCS histogram distance are as follows: 

For the RCS measurement $\mathcal{R^\prime}$,
we first normalize it to range $(0,1)$. To filter RCS of trivial objects, we only retain RCS within $(b_m, 1)$, where $b_m$ is the lower RCS bound and the bin width $b_w$ is empirically determined from a validation set.
Finally, by calculating counts in each bin, the RCS histogram is obtained.
For consistency across different scans, the histogram is normalized to have a unit sum over all bins.

Now the problem arises how to measure the similarity of two histograms. As a histogram is an empirical estimate of a probability distribution\cite{meshgi2015expanding}, we can use a wide range of similarity functions for comparing two distributions to measure the similarity of two histograms. In this work we use KL divergence as the similarity function since it is proven to be the an effective metric among a variety of similarity functions \cite{meshgi2015expanding}. By denoting RCS histogram distance of two RCS histograms $\widetilde{H}(\mathcal{R}^\prime_1)$ and $\widetilde{H}(\mathcal{R}^\prime_2)$ as $d_{R}$, we have:
\vspace{-0.15cm}
\begin{equation*}
    \label{dr}
d_{R}(\widetilde{H}(\mathcal{R}^\prime_1;k), \widetilde{H}(\mathcal{R}^\prime_2;k)) = \sum\limits_{k}\widetilde{H}(\mathcal{R}^\prime_1;k)\log(\frac{\widetilde{H}(\mathcal{R}^\prime_1;k)}{\widetilde{H}(\mathcal{R}^\prime_2;k)})
\vspace{-0.15cm}
\end{equation*}
where $k$ is the bin index. Note that a more similar RCS histogram pair leads to a smaller RCS histogram distance $d_{R}$. Algorithm ~\ref{rcs_alg} summarises the above procedures. With the derived RCS histograms and top-$M$ candidates retrieved based on their feature distance, a re-rank is performed by holistically considering the combined total distance $d_{total}$:
\begin{equation*}
    \label{d_total}
    d_{total}(q,c) = \alpha \cdot d_{R}(\widetilde{H}(q),\widetilde{H}(c)) + (1-\alpha) \cdot d_E(f(q), f(c))
\vspace{-0.1cm}
\end{equation*}
where $\alpha$ is an adjustable parameter used to balance the two distances, $q$ and $c$ denotes the query sample and the candidate sample. And we empirically set $M$=100 in this study. An example of RCSHR is shown in Fig. \ref{RCSHR} where we can see that the retrieved top-1 candidate without using the RCSHR is only plausibly correct but not really close to the query. After the RCSHR re-rank is applied, the real match pops up as the new top-1 candidate due to their more similar RCS histograms, and consequently smaller $d_{R}$ over the others.

\setlength{\textfloatsep}{0pt}
\begin{algorithm}[t]
    
    \label{rcs_alg}
    \LinesNotNumbered
    \SetKwInOut{Input}{Input}\SetKwInOut{Output}{Output}
    \caption{Calculate RCS Histogram Distance 
    }
    \Input{RCS measurements \( \mathcal{R}^\prime_1=\{r_{1,i}|i=1,2,...,N_1 \}\) \(\mathcal{R}^\prime_2=\{r_{2,i}|i=1,2,...,N_2 \}\), \(b_m\), \(b_w \)}
    \Output{RCS histogram distance \( d_{R}\)}
    
    \BlankLine
    bin the range \((b_m, 1)\) with equal width \(b_w\)\;
    \ForEach(){\(\mathcal{R^\prime}\)  \rm in \( \{\mathcal{R}^\prime_1, \mathcal{R}^\prime_2 \}\)}{
        \ForEach(){$r$ \rm in \(\mathcal{R^\prime}\)}{\(r \leftarrow \frac{r-\min{\mathcal{R^\prime}}}{\max{\mathcal{R^\prime}}-\min{\mathcal{R^\prime}}}\)}
        calculate counts in each bin and we get \(H(\mathcal{R^\prime})=\{ c_k|k=1,2,..[{b_m}/{b_w}]\}\)\;
    \( \widetilde{H}(\mathcal{R^\prime}) = \{\frac{c_k}{\sum\limits_{k}c_k}|k=1,2,..[{b_m}/{b_w}]\}\)\;
    }
    \BlankLine
    \(d_{R} = \sum\limits_{k}\widetilde{H}(\mathcal{R}^\prime_1;k)\log(\frac{\widetilde{H}(\mathcal{R}^\prime_1;k)}{\widetilde{H}(\mathcal{R}^\prime_2;k)}) \)

\end{algorithm}

\section{Experimental Setup}

\subsection{nuScenes Dataset}

As automotive radar is relatively new, 
the public dataset of this novel sensor is limited. The newly released nuScenes dataset\cite{caesarNuScenesMultimodalDataset2020} is the first and only one for large-scale environments with multi-modal sensors, including automotive radar. 
Since no previous work has been done for place recognition on this dataset, we provide necessary information about the dataset and our data pre-processing pipelines.

There are five radar sensors installed at the front, left, right and back parts of the vehicle, covering a $\SI{360}{\degree}$ FOV. Each radar works at $\SI{77}{\GHz}$ with $\SI{250}{m}$ measurement range and $\SI{13}{\Hz}$ capture frequency. The final data is comprised of 1000 scenes of $\SI{20}{\s}$ duration from four locations, which is $\SI{15}{\hour}$ of driving data ($\SI{242}{\km}$ traveled at an average of $\SI{16}{\km\per\hour}$).

In order to train and evaluate place recognition methods, a large number of loop closures and diverse road situations are desired. Therefore, we choose to train and evaluate \sysname on the largest split, \textit{Boston} split, among all the four splits. 
For the 550 scenes collected in Boston over 120 days, we take the data from the last 15 
days as the \textit{validation query set} and the \textit{test query set}, while the rest as the \textit{training set}. The \textit{training set} is further divided into the \textit{database set} and the \textit{training query set}. Specifically,
\begin{itemize}
    \item The \textit{database set} is created by taking as many places as possible from the \textit{training set}, until a newly added place is no $\SI{1}{\m}$ farther than any existing places in the \textit{database set}.
    \item The \textit{training query set} is created by substracting the \textit{database set} from the \textit{training set}.
    \item The \textit{validation query set} : the \textit{test query set}=1 : 4, and they are refined by removing places that have no ground truth true positives in the \textit{database set}.
\end{itemize}
The final \textit{database set}, \textit{training query set}, \textit{validation query set} and \textit{test query set} contains 6312, 7075, 924 and 3696 radar images, respectively.

\begin{figure}[t]
    \centering
    \includegraphics[width=2.5in]{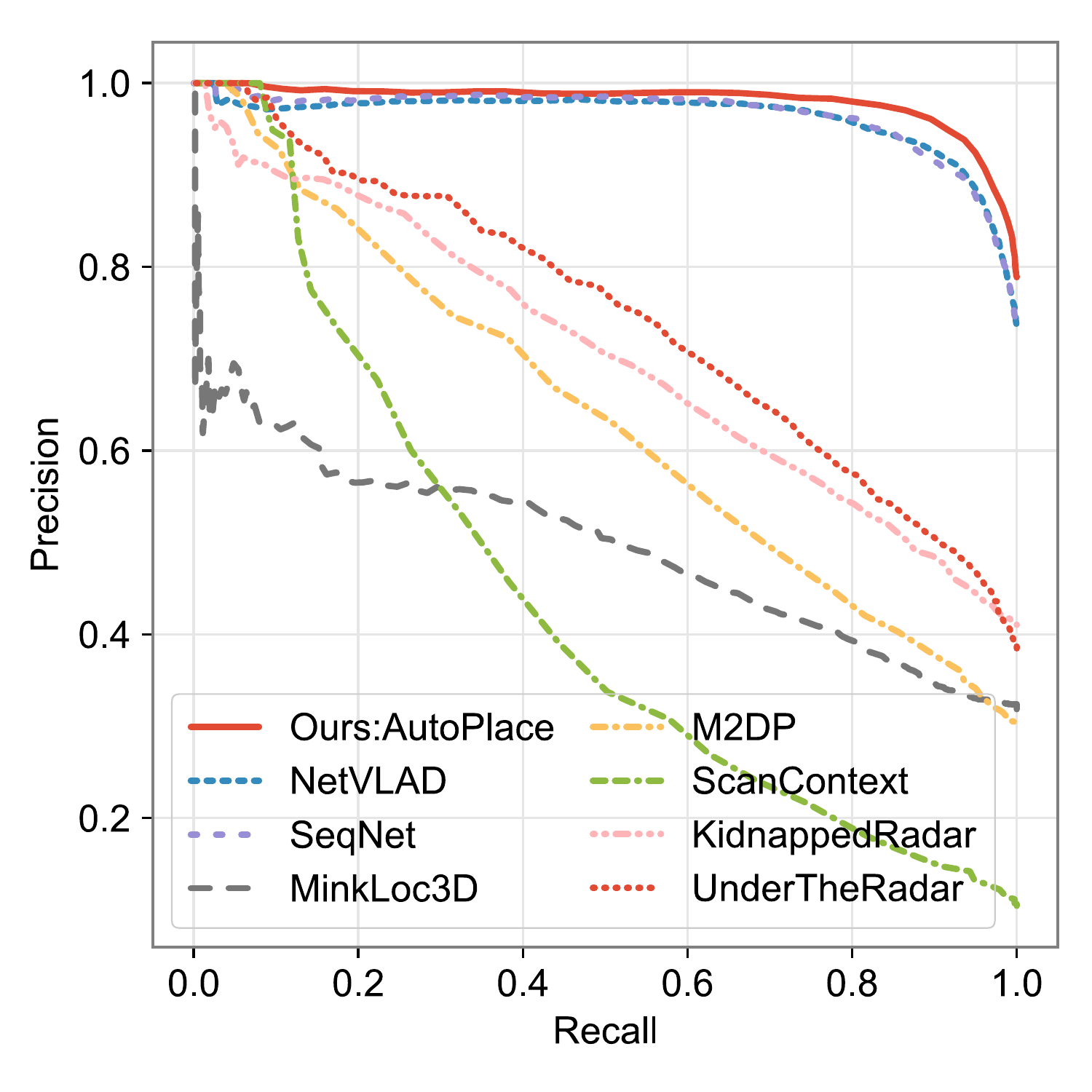}
    \setlength{\abovecaptionskip}{-0.2cm}

    \caption{Precision-recall curve of SOTA methods on the nuScenes dataset.}
    \label{pr-SOTA}
\vspace{0.2cm}
\end{figure}

\subsection{Implementation Details}

Since a single radar point cloud is too sparse to extract useful information due to the low sensing quality, we follow typical data pre-processing of \cite{liMillimeterWaveRadarSLAM2020,linDepthEstimationMonocular2020} and concatenate the nearest seven 
radar point clouds to form a denser point cloud. Ground truth ego-motion is used for concatenation, but this could be relaxed by using a simple local pose estimator, e.g., IMU. Besides, as the far-range measurement is less accurate, we crop the radar measurements to retain the points within $\SI{100}{m}$ from the sensor. When projected to the image plane, each point occupies one pixel. Thus, the converted 2D bird view radar image has a size of $200 \times 200$. 

After searching hyper-parameters on validation set, we set \(\tau=0.15, v_r^{\tau}=1, p^\tau=0.5\) for DPR, and \(b_m=0.02, b_w=0.04, \alpha=0.41 \) for RCSHR. For the network training, we use a batch size of 8 and SGD with an initial learning rate of $0.01$, momentum $0.9$ and weight decay $0.001$. We decay the learning rate by $0.5$ every $5$ epochs. 
Following the scale of KidnappedRadar\cite{saftescuKidnappedRadarTopological2020} and PointNetVLAD\cite{uyPointNetVLADDeepPoint2018}, we regard places in database that are within the radius=$\SI{9}{\m}$ area to the query as true positives, while those are outside the radius=$\SI{18}{\m}$ area as true negatives.

\subsection{Evaluation Metrics}

We follow four standard metrics in place recognition tasks: recall@N\cite{arandjelovicNetVLADCNNArchitecture2016,zhuAttentionbasedPyramidAggregation2018}, precision-recall curve\cite{houEvaluationObjectProposals2018, chenLearningContextFlexible2018}, $\max F_1$\cite{saftescuKidnappedRadarTopological2020} and average precision (AP)\cite{houEvaluationObjectProposals2018}. 
See \cite{houEvaluationObjectProposals2018} for details of generating precision-recall curve and AP.

\section{Results}

\subsection{Comparison with State-Of-The-Art Methods}

In what follows, we denote our proposed Spatial Encoder and Temporal Encoder as \textbf{\texttt{SE}} and \textbf{\texttt{TE}} for brevity. We compare our approach with the SOTA methods, including:
\begin{itemize}
    \item Visual place recognition: \textbf{NetVLAD}\cite{arandjelovicNetVLADCNNArchitecture2016}, \textbf{SeqNet}\cite{gargSeqNetLearningDescriptors2021}.
    We adapt the implementation of the above works to the settings of the nuScenes dataset. To fairly evaluate the effectiveness of our spatial encoder, we also investigated the performance of \textbf{NetVLAD}+\textbf{\texttt{TE}} by adding our \texttt{TE} to the original NetVLAD network.
    \item LiDAR place recognition: \textbf{M2DP}\cite{heM2DPNovel3D2016}, \textbf{ScanContext}\cite{kimScanContextEgocentric2018} and \textbf{MinkLoc3D}\cite{komorowskiMinkLoc3DPointCloud2020}. We feed them the pseudo-3D point clouds by adding a pseudo axis $z=0$ and then normalize point clouds to range $(-1,1)$.
    \item Spinning radar place recognition: \textbf{UnderTheRadar}\cite{barnesRadarLearningPredict2020} and \textbf{KidnappedRadar}\cite{saftescuKidnappedRadarTopological2020}. For KidnappedRadar, we convert the radar images from Cartesian coordinates to polar coordinates. 
\end{itemize}

\begin{table}
    \vspace{0.15cm}
    \centering
    \caption{Performance of SOTA methods on the nuScenes Dataset}
    \label{recall-table-SOTA}
    \begin{threeparttable}

    \begin{tabular}{lccc} 
    \hline
    Method          & Recall@1/5/10                                                                                                              & $\max F_{1}$  & AP             \\ 
    \hline
    MinkLoc3D\cite{komorowskiMinkLoc3DPointCloud2020}       & 31.8/53.6/61.1 & 0.53 & 0.49           \\
    M2DP\cite{heM2DPNovel3D2016}            & 30.3/43.4/48.4 & 0.58 & 0.65           \\
    ScanContext\cite{kimScanContextEgocentric2018}     & 10.4/15.3/17.2 & 0.42 & 0.45           \\ 
    \hline
    UnderTheRadar\cite{barnesRadarLearningPredict2020} &38.5/53.8/59.1 & 0.67 & 0.75           \\
    KidnappedRadar\cite{saftescuKidnappedRadarTopological2020} & 41.0/56.6/61.5 & 0.65 & 0.71           \\ 
    \hline
    NetVLAD\cite{arandjelovicNetVLADCNNArchitecture2016}         & 73.1/80.5/82.4 & 0.92 & 0.96           \\
    NetVLAD+\texttt{TE}*\tnote{1}  & 70.8/79.5/81.2 & 0.90 & 0.95           \\
    SeqNet\cite{gargSeqNetLearningDescriptors2021}*          & 73.3/80.0/82.1 & 0.92 & 0.97           \\ 
    \hline
    Ours: \texttt{SE}        & 73.3/80.0/81.9 & 0.89 & 0.96           \\    
    Ours: \texttt{SE+TE}*        & 76.7/81.7/83.4 & 0.93 & 0.97           \\    
    Ours: \sysname*\tnote{2}       & \textbf{78.9/83.1/84.3} & \textbf{0.94} & \textbf{0.98}  \\
    \hline
    \end{tabular}
    \begin{tablenotes}
        \footnotesize
        \item[1] * denotes using sequential frames for place recognition.
        \item[2] i.e., \texttt{SE+TE+DPR+RCSHR}.
      \end{tablenotes}
    \end{threeparttable}     
\end{table}

\label{experiments-sota}

\begin{table}
    \centering
    \caption{Ablation Study of \sysname}
    \label{score-ablation-full}
    \begin{tabular}
    {cccc|ccc} 
    \hline
    \begin{tabular}[c]{@{}c@{}}\texttt{SE}\end{tabular} &\texttt{TE}      & \texttt{DPR}    & \texttt{RCSHR}  & Recall@1/5/10  & $\max F_{1}$ & AP    \\ 
    \hline
    $\surd$ &         &         &           & 73.3/80.0/81.9 & 0.89 & 0.96  \\
    \hline
    $\surd$ & $\surd$ &         &           & 76.7/81.7/83.4 & 0.93 & 0.97  \\ 
    \hline
    $\surd$ &         & $\surd$ &           & 75.3/81.4/83.0 & 0.91 & 0.96 \\ 
    \hline
    $\surd$ &         &         &  $\surd$  & 73.4/80.3/82.2 & 0.89 & 0.96  \\ 

    \hline
    $\surd$ &         & $\surd$ & $\surd$ & 75.8/82.1/83.8 & 0.91 & 0.96  \\ 
    \hline
    $\surd$ & $\surd$ &         & $\surd$ & 77.7/82.1/83.4 & 0.93 & 0.98  \\ 
    \hline
    $\surd$ & $\surd$ & $\surd$ &         & 77.8/82.3/83.7 & 0.94 & 0.98  \\ 
    \hline    
    $\surd$ & $\surd$ & $\surd$ & $\surd$ & \textbf{78.9}/\textbf{83.1}/\textbf{84.3} & \textbf{0.94} & \textbf{0.98} \\
    \hline
    \end{tabular}
\end{table}

\begin{figure*}[htbp!]
    \centering
    \vspace{0.15cm}
    \setlength{\abovecaptionskip}{0.cm}
    \includegraphics[width=\textwidth]{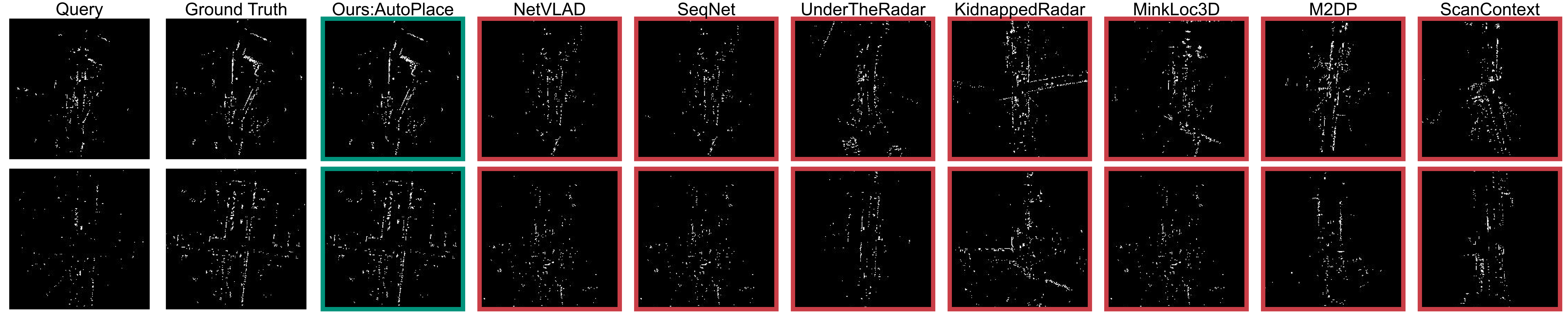}
    \caption{Qualitative analysis of SOTA methods. The first and second columns show the query radar images and ground truth radar images, and the other columns are the retrieved top 1 candidate via different methods. \textcolor{RightColor}{Green} means the retrived candidate is a true positive, while \textcolor{WrongColor}{red} denotes false positive.
    } 
    \label{qua-SOTA}
\vspace{-0.5cm}
\end{figure*}

Table. \ref{recall-table-SOTA} presents the performance of SOTA methods and our methods. As expected
, brute-force applying the place recognition approaches of LiDAR sensors to automotive radar results in inferior performance: ScanContext, M2DP and MinkLoc3D only achieve 10.4\%, 30.3\% and 31.8\% recall@1, respectively. 
Their failures can be attributed to the reasons that (1) automotive radar point clouds are much sparser than LiDAR point clouds\cite{linDepthEstimationMonocular2020}, and (2) pseudo 3D point clouds of the automotive radar lack valid information on $z$ axis. These factors make 3D point cloud-based methods ill-suited to automotive radar.

We can also observe from this table that spinning radar place recognition approaches perform slightly better than LiDAR-based methods: KidnappedRadar and UnderTheRadar achieve 41.0\% and 38.5\% recall@1, respectively, which are still far from being satisfactory. KidnappedRadar takes mechanically spinning radar's spectra as input and performs max-pooling upon the last feature map along the azimuth axis to achieve rotational invariance. However, since automotive radar point clouds are higher-level but less informative than spectra, max-pooling operation on automotive radar's feature map only makes things worse, preventing the network from producing a discriminative radar image descriptor. 

Visual place recognition methods, NetVLAD and SeqNet, achieve 73.1\% and 73.3\% recall@1, respectively. We suppose their relatively good performance results from the strong feature representative capability of VGG16\cite{simonyan2014very}. 
Our \texttt{SE} surpasses NetVLAD and achieves a comparable performance of the sequence-based method, SeqNet.  
Notice that NetVLAD+\texttt{TE} performs even worse than NetVLAD, this is because NetVLAD is designed to produce high dimensional features (larger than 30k-dimensions in \cite{arandjelovicNetVLADCNNArchitecture2016}) while LSTM in \texttt{TE} works well when fed low dimensional data (less than 10k-dimensions in \cite{wangEndtoendSequencetosequenceProbabilistic2018, luMilliEgoSinglechipMmWave2020}), such a incompatibility results in its poor performance. In contrast, our \texttt{SE} produces compact features and works better with \texttt{TE} than NetVLAD.
By utilizing all information from automotive radar, 
\sysname  extends the gap to the runner-up to 5.6\%, 0.02 and 0.01 for recall@1, $\max F_1$ and AP.
A similar trend can also be observed in Fig. \ref{pr-SOTA} that LiDAR and spinning radar place recognition methods are outperformed by their visual counterparts, NetVLAD and SeqNet.
Still, \sysname exceeds the others by a significant margin.

We also provide qualitative analysis in Fig. \ref{qua-SOTA}. As we can see, when the queried scene structure is incomplete (first row) or the point cloud in a query is extremely sparse (second row), competing approaches struggle while \sysname can still retrieve the correct match.

In summary, the  experimental results suggest:
 (1) existing visual, LiDAR or spinning radar place recognition methods perform unsatisfactorily when being directly applied to automotive radar, and (2) by fully utilizing all the information provided by automotive radar, our \sysname is superior to all competing methods.

\subsection{Ablation Study}




We study each component of \sysname by evaluating different groups shown in Table. \ref{score-ablation-full}. It can be observed that 
\begin{itemize}
    \item \textbf{SE} (c.f. Sec.~\ref{ssub:se}) alone achieves recall@1=73.3\%, which is comparable with NetVLAD.
    \item \textbf{TE} (c.f. Sec.~\ref{ssub:te}) boosts recall@1 by 3.4\% when added to \texttt{SE}, and decreases it by 3.1\% when removed from \sysname.
    \item\textbf{DPR} (c.f. Sec.~\ref{ssub:dpr}) increases recall@1 by 2.0\% when added to \texttt{SE}, and decreases it by 1.2\% when removed from \sysname.
    \item \textbf{RCSHR} (c.f. Sec. \ref{subs:RCSHR}) improves recall@1 by 0.1\%, 0.5\%, 1.0\% and 1.1\% when added to \texttt{SE}, \texttt{SE+DPR}, \texttt{SE+TE} and \texttt{SE+TE+DPR}, respectively. Improvements vary because \texttt{RCSHR} works as a refinement module, bringing more remarkable improvement when the descriptors themselves are more discriminative (i.e., \texttt{RCSHR} hardly changes candidates' order when mismatches have small feature distances). This is in accordance with our observation that more discriminative descriptors lead to fewer mismatches between geometrically-close places, and thus, \texttt{RCSHR} helps more in distinguishing these places
    (more illustrative examples are available in our supplementary video).
\end{itemize}
The results indicate that each component is critical in improving the overall performance, of which the most significant benefit is from \texttt{TE}.

\section{Conclusion}
Observing that existing place recognition methods ill-suits the emerging automotive radar, we propose \sysname, a novel place recognition framework by fully exploiting automotive radar's rich measurements. Experimental results show remarkable performance gain of \sysname on the public nuScenes dataset. Future work will further investigate \sysname for simultaneous localization and mapping.

\bibliographystyle{IEEEtran}
\bibliography{MyLibrary}
\end{document}